\documentclass[10pt,twocolumn,letterpaper]{article}

\usepackage{cvpr}
\usepackage{times}
\usepackage{epsfig}
\usepackage{graphicx}
\usepackage{amsmath}
\usepackage{amssymb}

\usepackage{subfigure}
\usepackage{comment}
\usepackage{color}
\usepackage{ulem}
\usepackage{url}
\usepackage{newtxtext,newtxmath}

\newcommand{\cHiro}[1]{{\color{black}{#1}}}

\newcommand{\cEpf}[1]{{\color{black}{#1}}}

\makeatletter
\newcommand{\figcaption}[1]{\def\@captype{figure}\caption{#1}}
\newcommand{\tblcaption}[1]{\def\@captype{table}\caption{#1}}
\makeatother

\usepackage[breaklinks=true,bookmarks=false]{hyperref}

\cvprfinalcopy 

\def\cvprPaperID{****} 

\begin{document}

\title{Weakly Supervised Dataset Collection\\for Robust Person Detection}

\author{Munetaka Minoguchi, Ken Okayama, Yutaka Satoh, Hirokatsu Kataoka\\
National Institute of Advanced Industrial Science and Technology (AIST)\\
Tsukuba, Ibaraki, Japan\\
}

\maketitle

\begin{abstract}
   To construct an algorithm that can provide robust person detection, we present a dataset with over 8 million images that was produced in a weakly supervised manner. Through labor-intensive human annotation, the person detection research community has produced relatively small datasets containing on the order of 100,000 images, such as the EuroCity Persons dataset, which includes 240,000 bounding boxes. Therefore, we have collected 8.7 million images of persons based on a two-step collection process, namely person detection with an existing detector and data refinement for false positive suppression. According to the experimental results, the Weakly Supervised Person Dataset (WSPD)  is simple yet effective for person detection pre-training. In the context of pre-trained person detection algorithms, our WSPD pre-trained model has 13.38 and 6.38\% better accuracy than the same model trained on the fully supervised ImageNet and EuroCity Persons datasets, respectively, when verified with the Caltech Pedestrian. The dataset and pre-trained models used in the paper are publicly available on the GitHub\footnote{\url{https://github.com/cvpaperchallenge/FashionCultureDataBase_DLoader}}. The paper is under consideration at Pattern Recognition Letters.
\end{abstract}

\section{Introduction}




In the context of navigation and service robots, an appropriate human-centered operating environment usually starts with person detection. Therefore, we require robust and highly accurate person detection for collision avoidance for the realization of self-driving cars, mobile robots, and unmanned aerial vehicles.

To construct a learning-based object detector
(In this paper, the meaning of ``object'' detection includes person detection; in other words, ``object'' is taken as having a broader meaning)
, a large-scale and well-labeled dataset is required, as is a suitable model architecture. Towards this end, large-scale multiple-object datasets, such as MS COCO~\cite{refs:MSCOCO} and OpenImages~\cite{refs:OpenImages}, have been produced to conduct highly accurate object detection. However, the datasets for person detection are currently relatively small. For example, the Caltech Pedestrian~\cite{refs:CaltechPeds}, CityPersons~\cite{refs:CityPersons}, and EuroCity Persons datasets~\cite{refs:braun2019eurocity} contain 350,000, 35,000, and 240,000 bounding boxes (bboxes), respectively. Compared to million-image multiple-object datasets like OpenImages, pedestrian datasets could still be greatly expanded to improve the performance of models that are trained with these datasets. Thus, there is great motivation to create a large-scale person dataset with a bboxes quantity on the order of millions.

Inspired by weakly supervised image labeling with Social Network Service (SNS) hashtags, as in the so-called ``Instagram-3.5B'' study~\cite{refs:Instagram3.5B},  we conducted semi-automatic large-scale dataset collection in the context of person detection. Unlike the related work in Instagram-3.5B, our proposed dataset contains a large number of bboxes in addition to the captured images. We here consider a different method for constructing a pre-trained person dataset based on an existing detector and simple refinement.

\begin{figure*}[t]
\begin{center}
\includegraphics[width=1.0\linewidth]{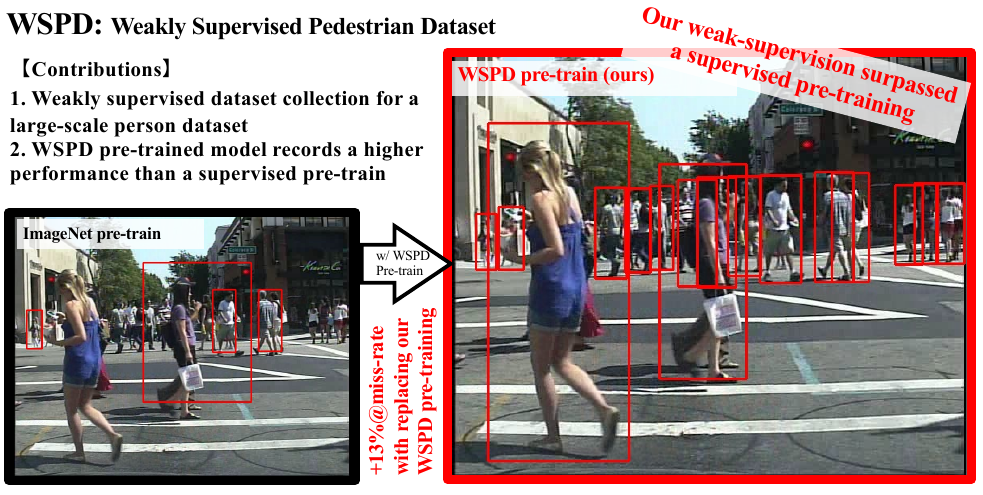}
\end{center}
   \vspace{0.0pt}\caption{\cHiro{Overview of the proposed Weakly Supervised Person Dataset (WSPD) and its contributions.}}
\label{fig:wspd}
\end{figure*} 

This paper proposes a million-image person dataset based on a weakly supervised method for robust person detection, namely Weakly Supervised Person Detection (WSPD). Our large-scale person dataset is constructed by semi-automatic image collection and data refinement with SNS images. This dataset collection method allows us to significantly improve the \cHiro{performance of person detection with human annotation in a few hours}. When used as a pre-trained person dataset, the model trained with the WSPD method outperforms the detection rate of models trained with fully supervised pre-trained datasets, such as EuroCity Persons (+6.38\%) and ImageNet (+13.36\%) on the Caltech Pedestrian with a Single-Shot multibox Detector (SSD)~\cite{refs:SSD}. 

\begin{figure*}[t]
\begin{center}
\includegraphics[width=1.0\linewidth]{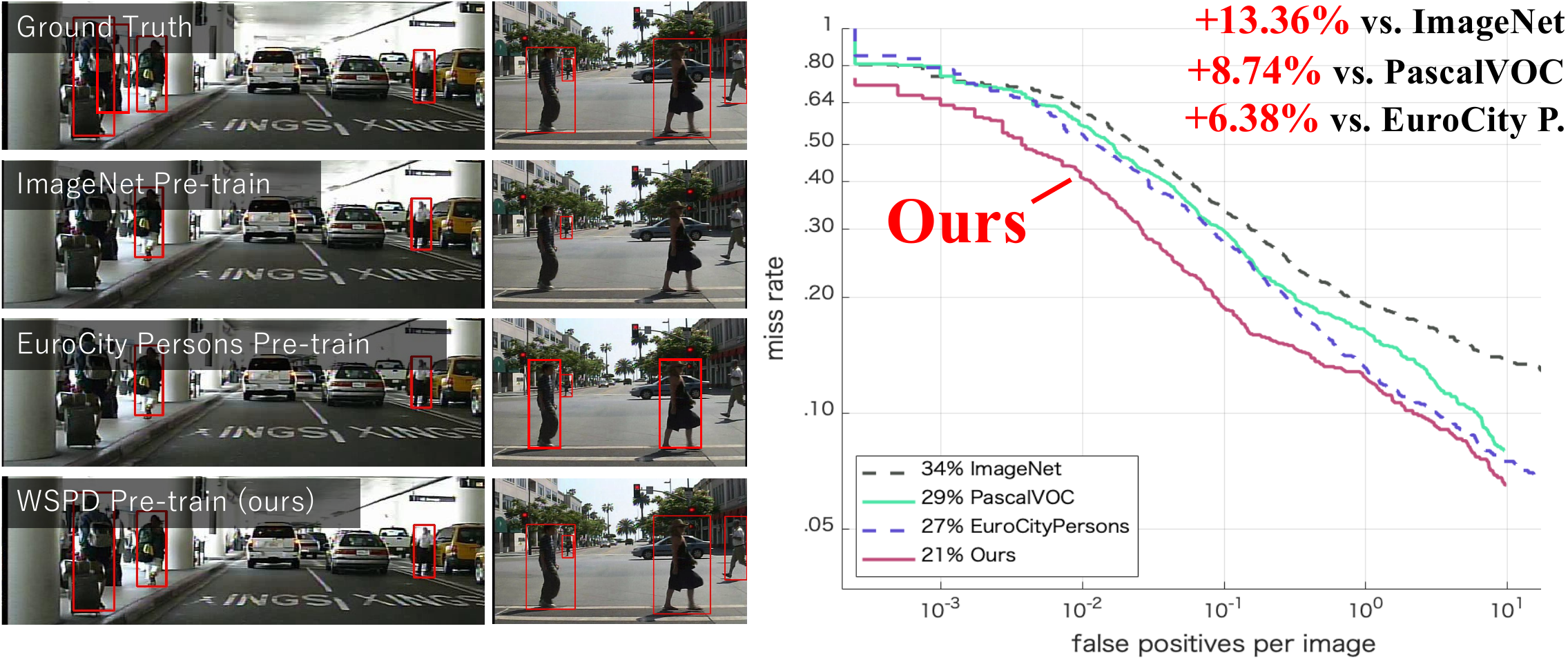}
\end{center}
   \vspace{0.0pt}\caption{\cEpf{We have constructed a million-image person dataset for use with pre-trained person detectors. (Left) Our WSPD method, which creates a large-scale pre-trained person dataset, provides better person detection performance. We list examples of ground truth and the detection results for the ImageNet and EuroCity Persons pre-trained models. (Right) Detection error trade-off (DET) curves for the baseline models (ImageNet, Pascal VOC, and EuroCity Persons) versus the proposed model. The miss rate (\%) for the Caltech Pedestrian is shown, with higher values representing a higher accuracy, that is, our proposed method is up to 13.36\% better than the baseline. The baselines employ the ImageNet/Pascal VOC/EuroCity Persons pre-trained VGG-16 neural network and Caltech Pedestrian fine-tuned SSD. In our proposed method, we use the WSPD pre-trained model and Caltech Pedestrian fine-tuned SSD. Note that our proposed WSPD, Pascal VOC, and EuroCity Persons pre-trained models solved person detection tasks with both the pre-trained and fine-tuned datasets.}}    
\label{fig:results}
\end{figure*}

This paper makes the following contributions to person detection \cHiro{(see also Figure~\ref{fig:wspd})}. (i) We propose a weakly supervised dataset collection method for building training datasets for person detection. We then use this method to construct a dataset containing millions of images with bboxes through an existing detector (e.g., Faster R-CNN~\cite{refs:FasterR-CNN}) and false positive suppression. (ii) The WSPD pre-trained model is demonstrated to perform well at person detection. The fine-tuned detector achieved  an accuracy 6.38 and 13.36\%  higher than that of the baseline models (EuroCity Persons and ImageNet, respectively) on the Caltech Pedestrian. \cHiro{We provide examples of the detection results and performance comparisons in Figure~\ref{fig:results}.}

\section{Related work}
\cEpf{In this section we briefly review some of the key concepts related to this paper, such as object detection, annotation, and dataset collection, to help highlight how the proposed method differs from existing methods.}

\textbf{Object detection.} Object detection algorithms have progressed from hand-crafted detection using local features (e.g., Haar-like features~\cite{refs:Haar-like}, HOG~\cite{refs:HOG}, and ICF~\cite{refs:ICF}) and well-organized classifiers (e.g., Deformable Parts Model~\cite{refs:FelzenszwalbTPAMI2010} and aggregated detectors~\cite{refs:acf}), and currently we are in the era of deep neural networks (DNNs). In the literature, a two-step region identifier and DNN-based classifier has been proposed~\cite{refs:R-CNN}. The basic technique, called R-CNN, has been adapted for use with any-size feature maps~\cite{refs:GirshickICCV2015}, and it includes an end-to-end two-step method~\cite{refs:FasterR-CNN}. Current research is widely divided between one-shot detectors, such as you only look once (YOLO)~\cite{refs:YOLO} and SSD~\cite{refs:SSD}. Recent studies have also focused on highly accurate detectors, such as RetinaNet~\cite{refs:Retina} and M2Det~\cite{refs:M2Det}, and instance segmentation with Mask R-CNN~\cite{refs:Mask}. Here, we use SSD, which is a balanced detector with a relatively short training time. Further, it can easily be optimized for use with baseline models to compare the dataset prepared with our collection method and the ImageNet and EuroCity Persons datasets in the context of a pre-trained detector. Moreover, we implemented M2Det to identify which person detector is more accurate.

\textbf{Person detection.} \cEpf{According to a comprehensive survey~\cite{refs:BenensonECCVW2014}, the performance rate of person detection algorithms has increased over the last decade as person detectors have evolved to use more sophisticated architectures. A recent study has proposed several configurations to improve recognition and localization with DNNs~\cite{refs:HosangCVPR2015}, semantic meaning~\cite{refs:DuWACV2017}, combined methods~\cite{refs:ZhangECCV2016}, and analysis of small images or crowds~\cite{refs:WangCVPR2018_repulsion}.}
However, a large-scale person dataset must be prepared for training these models (e.g., SSD or M2Det) and fine-tuning their architecture.

\textbf{Annotation for object bboxes.} In recent machine learning research, annotation treatment has been shown to be important for successful network training. For example, Su~\textit{et al.} introduced a method for repeatedly checking bbox annotations in an image in three steps: drawing, quality checking, and coverage verification~\cite{refs:SuAAAIW2012}. Papadopoulos~\textit{et al.} proposed a method that combines an existing detector and human annotation~\cite{refs:Papadopoulos}. The combined method iterates between three annotation and quality control steps: model retraining, bbox relocalization, and human verification. Their annotation method results in an object detection dataset without human-drawn bboxes. Compared with these annotation methods, our proposed dataset collection method requires only a minimum of human-based annotation checks to improve the performance of person detection.

\textbf{Person dataset collection.} In addition to changes in the models used for person detection, person datasets have also evolved over the last decade. The first generation of person detection datasets consisted of small training and testing datasets (up to 10,000 images, including INRIA~\cite{refs:HOG}, Daimler~\cite{refs:daimler}, and ETHZ~\cite{refs:ethz}), followed by a second generation of medium-size datasets (10,000--100,000 images, including Caltech Pedestrian~\cite{refs:CaltechPeds}, CityPersons~\cite{refs:CityPersons}, and EuroCity Persons~\cite{refs:braun2019eurocity}) that include occlusion and cluttered backgrounds. However, to the best of our knowledge, a large-scale person dataset (over 1 million images) is not currently freely available. In \cite{refs:ZhangCVPR2016}, it was claimed that more high-quality person annotations are required for improving the accuracy of person detection algorithms. Data collection with weak supervision is one area of ongoing work in image classification~\cite{refs:Instagram3.5B}. As discussed in related work~\cite{refs:JFT300M}, the increasing scale of datasets is also important for improving the accuracy of existing detection algorithms. To help produce large-scale datasets, we here present a weakly supervised pre-training dataset annotation method for person detection.

\begin{figure*}
\begin{center}
\includegraphics[width=1.0\linewidth]{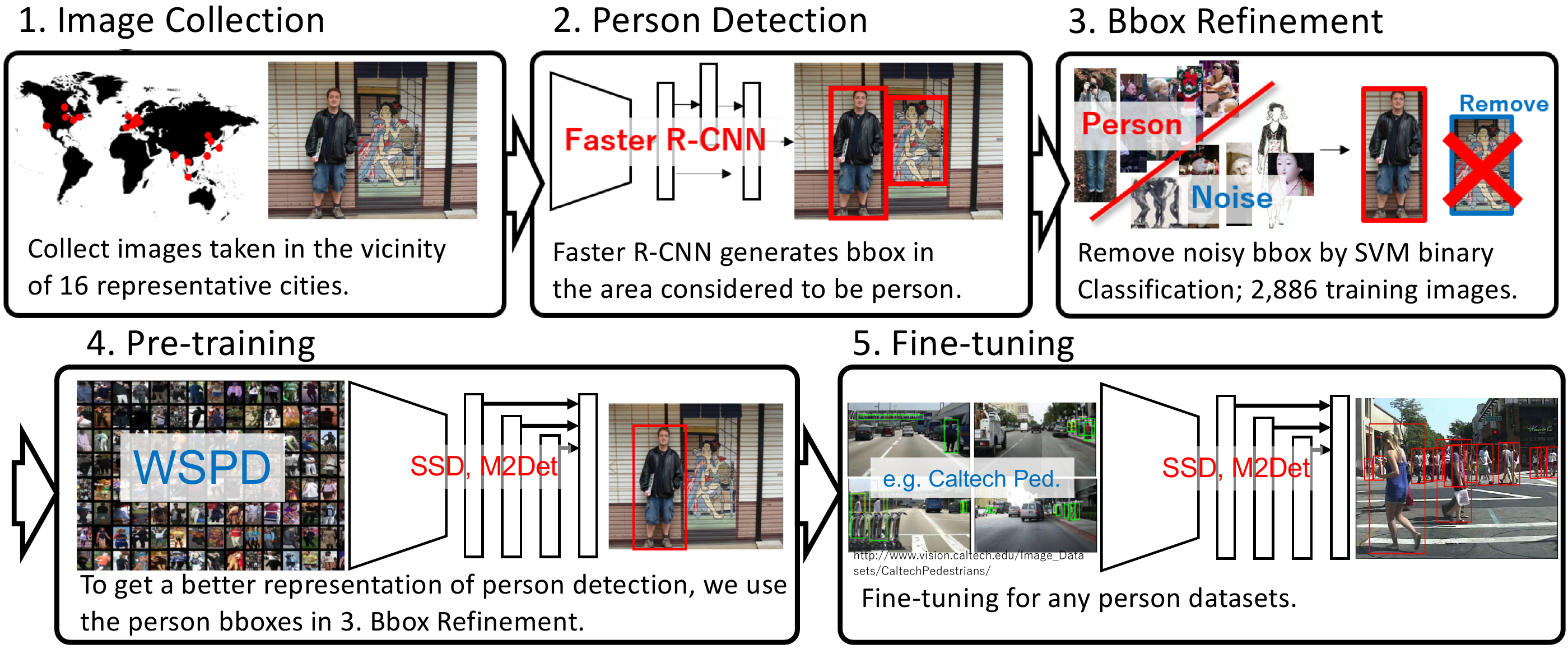}
\end{center}
   \vspace{0.0pt}\caption{\cEpf{Weakly supervised dataset collection. (1)  Cloud-based image download and collection. Although this paper used the YFCC100M dataset intended for city perception~\cite{refs:ZhouECCV2014}, any image dataset can be used. (2) Detection of people with Faster R-CNN to add bboxes to the selected images. (3) Data refinement to exclude unwanted bboxes with a binary classifier, which determines whether a person's whole body is contained within the bbox.}}
\label{fig:concept}
\end{figure*} 

\section{Weakly supervised dataset collection}
\subsection{Overview}

To obtain a better representation of persons for detection during pre-training, a large-scale dataset with bboxes should be used, such as a combination of our WSPD pre-trained model and the Caltech Pedestrian~\cite{refs:CaltechPeds} fine-tuned detector. The image localization labeling depends on the efforts of human annotators; therefore, an automatic dataset creation method would be useful for the person detection research community. 


Figure~\ref{fig:concept} illustrates the concept of weakly supervised dataset collection. After collecting a large number of SNS images, we apply a two-step algorithm for weakly supervised dataset construction: person detection with an existing object detector and erasing false positives using a binary classifier. \cHiro{Here, we describe the problem setting to conduct our weakly supervised dataset collection. The setting of weakly supervised learning is simple yet effective for pre-training a person detector. At the beginning, we assign an object detector $D$ to recognize bboxes and their labels from an input image $x$.}
\begin{eqnarray}
   (y^{'}, b_{box}^{'}) &=& D(x; \theta),
\end{eqnarray}
\cHiro{where $y^{'}$ and $b^{'}$ denote predictions of object category and bboxes, respectively, and $\theta$ represents trained parameters in the detector. In case of person detection, the category is limited to the ``person'' label. The equation is simplified as follows:}
\begin{eqnarray}
   b_{box}^{'} &=& D(x; \theta).
\end{eqnarray}
\cHiro{We used a support vector machine (SVM) to refine the detected bboxes with $D$. We used weakly supervised dataset collection to classify the detected bboxes as person ground truth ($y_{gt}$) or background ($y_{bg}$). The following equation shows the binary classifier with SVM:}
\begin{eqnarray}
   f(x) &=& w^{T}I(D(x; \theta)) + b,
\end{eqnarray}
\cHiro{where $I(*)$ represents a cropped image with detected bbox. The cropped image is divided depending on the $f(x)$.}
\begin{eqnarray}
  g(x) = \begin{cases}
    y_{gt} & (f(x) \ge 0) \\
    y_{bg} & (f(x) < 0)
  \end{cases},
\end{eqnarray}
\cHiro{where $y_{gt}$ is assigned to add a person label in the WSPD method.} We use the Faster R-CNN as a person detector~\cite{refs:FasterR-CNN} and a binary classifier for determining whether the target's whole body is contained within the bbox. Our framework is simple yet effective for generating a large-scale dataset. We use the Yahoo! Creative Commons 100M Database (YFCC100M)~\cite{refs:BartCVPR2016}, which contains 100 million Flickr images. Our person dataset (WSPD) contains images of people from around the world but is limited to specific major cities~\cite{refs:ZhouECCV2014}. The dataset consists of 2,822,421 original images and 8,716,461 person images (in bboxes). To the best of our knowledge, this is the largest person dataset for bbox-based detection currently available (see Table~\ref{tab:ourdb}).



\subsection{Data collection, refinement, and configuration}

\textbf{Collection.} We downloaded images from 21 global cities based on~\cite{refs:ZhouECCV2014}; however, we excluded cities having fewer than 100,000 collected images. Consequently, 16 of the 21 cities were selected for the WSPD: London, New York, Boston, Paris, Toronto, Barcelona, Tokyo, San Francisco, Hong Kong, Zurich, Seoul, Beijing, Bangkok, Singapore, Kuala Lumpur, and New Delhi (listed from most images to fewest images).  These metropolitan areas do not overlap, as they are at least 200 km apart.  To create the images with bboxes, we applied the Pascal VOC pre-trained VGG16 model for the Faster R-CNN.
We initially set the threshold value as 0.8 and used only the person label.
A dataset consisting of a geo-tag and a time-stamp was replicated from the YFCC100M dataset. 
In the first step, we collected 76,532,519 images using automatic image collection and bbox annotation.

\textbf{Refinement.} We now consider how to exclude noisy images from our dataset.
The refinement strategy is to scan all images with a simple classifier based on a combination of StyleNet~\cite{refs:SimoSerraCVPR2016} and a SVM.
To create a sophisticated fashion-oriented database, we treat the WSPD refinement as a binary classification problem to distinguish between street-fashion-snapshot whole-body images and other cropped images, such as partial bodies or backgrounds without a person.
We trained and refined the database with 1,443 carefully annotated objective images and a large number of randomly cropped negative images.

\noindent{\textbf{Configuration.} 
Our WSPD has the three following features:}
\begin{itemize}
    \item Images captured from the YFCC100M dataset. After data refinement, the number of images decreased from 8,504037 to 2,822,421 original images.
    \item \cEpf{Cropped person images with bboxes (treated as clothing images). As a result of data refinement, the number of person bboxes was reduced from 76,532,519 to 8,716,461.}
    \item Geo-location and time-stamp information. This relates to the 16 cities listed above.
\end{itemize}


\begin{table}
\begin{center}
\begin{tabular}{llrcc}
\hline
\multicolumn{1}{l}{Database} & \multicolumn{1}{r}{\#image} & \multicolumn{1}{c}{\#bbox} & \multicolumn{1}{c}{\#class}\\
\hline
\multicolumn{1}{l}{Pascal VOC~\cite{refs:EveringhamIJCV15}} &  \multicolumn{1}{r}{11,530} & \multicolumn{1}{r}{27,450} & \multicolumn{1}{r}{20}\\

\multicolumn{1}{l}{MS COCO~\cite{refs:MSCOCO}} & \multicolumn{1}{r}{123,287} & \multicolumn{1}{r}{ 896,782} & \multicolumn{1}{r}{80}\\

\multicolumn{1}{l}{OpenImages V5~\cite{refs:OpenImages}} & \multicolumn{1}{r}{1,743,042} & \multicolumn{1}{r}{14,610,229} & \multicolumn{1}{r}{600}\\
\hline
\multicolumn{1}{l}{Caltech Pedestrian~\cite{refs:CaltechPeds}} & \multicolumn{1}{r}{250,000} & \multicolumn{1}{r}{350,000} & \multicolumn{1}{r}{2}\\

\multicolumn{1}{l}{CityPersons~\cite{refs:CityPersons}} & \multicolumn{1}{r}{5,000} & \multicolumn{1}{r}{35,016} & \multicolumn{1}{r}{2}\\

\multicolumn{1}{l}{EuroCity Persons~\cite{refs:braun2019eurocity}} & \multicolumn{1}{r}{47,300} & \multicolumn{1}{r}{238,200} & \multicolumn{1}{r}{17}\\

\multicolumn{1}{l}{WSPD (proposed)} & \multicolumn{1}{r}{\textbf{2,822,421}} & \multicolumn{1}{r}{\textbf{8,716,461}} & \multicolumn{1}{r}{2}\\
\hline
\end{tabular}
\end{center}
\vspace{0pt}\caption{Proposed WSPD and related datasets.}
\label{tab:ourdb}
\end{table}

\textbf{Details of datasets.} We give details on the self-collected dataset in Table~\ref{tab:ourdb}. We also compare the proposed dataset (WSPD) with existing datasets for object and person detection. It is clear that our dataset contains the largest number of images and bboxes among the currently available person datasets. Also, our dataset contains a diversity of person images from different locations worldwide. From the semi-automatic dataset collection, we obtained millions of person bboxes that can be useful for training and testing a pre-trained detector.

\begin{figure}[t]
\begin{center}
\includegraphics[width=1.0\linewidth]{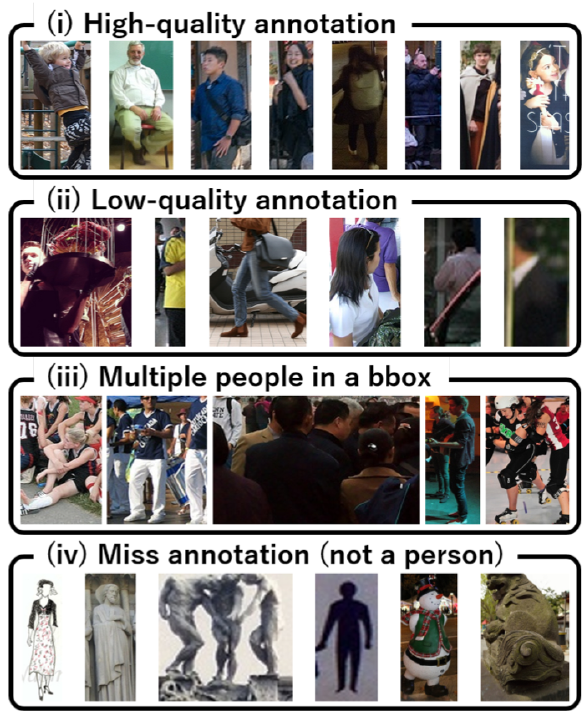}
\end{center}
   \vspace{0pt}\caption{Data quality analysis.}
\label{fig:datasetquality}
\end{figure}

\textbf{Dataset quality analysis.} We manually analyzed the results of our method using 1,000 randomly selected person bboxes from the WSPD dataset. Figure~\ref{fig:datasetquality} shows the four classifications for the randomly selected images, and Table~\ref{tab:dataquality} indicates the corresponding frequency of occurrence of each class. The four bbox classifications were (i) high-quality annotation, (ii) low-quality annotation (partial image of a person), (iii) multiple persons in a bbox, and (iv) misclassification (not a person). Based on our random sample of 1,000 images, we expect the collected dataset to consist of 93\% person images (i, ii, and iii combined), regardless of whether the images are perfectly annotated. Non-person images account for only 70 out of 1,000 bboxes. According to the Instagram-3.5B paper~\cite{refs:Instagram3.5B}, 10 and 25\% noise reduced the performance rate by only 1.0 and 2.0\%, respectively.

The effectiveness of the proposed weakly supervised dataset collection is shown in Section 5. We have considered multiple pre-trained datasets. 

\begin{table}
\begin{center}
\begin{tabular}{lc}
\hline
\multicolumn{1}{l}{Type of bbox annotation} & \multicolumn{1}{c}{\%}\\
\hline
\multicolumn{1}{l}{(i) High-quality annotation} & \multicolumn{1}{c}{62.2}\\
\multicolumn{1}{l}{(ii) Low-quality annotation} & \multicolumn{1}{c}{21.1}\\
\multicolumn{1}{l}{(iii) Multiple people in a bbox} & \multicolumn{1}{c}{9.7}\\
\multicolumn{1}{l}{(iv) Misclassification (not a person)} & \multicolumn{1}{c}{7.0}\\
\hline
\end{tabular}
\end{center}
\vspace{0pt}\caption{Statistics of dataset quality analysis. We randomly selected 1,000 images from the WSPD dataset.}
\label{tab:dataquality}
\end{table}

\begin{table*}
\begin{center}
\begin{tabular}{lccc}
\hline
\multicolumn{1}{l}{Method} & \multicolumn{1}{l}{Annotation} &  \multicolumn{1}{c}{Pre-training (\#classes, \#images)} &  \multicolumn{1}{c}{Fine-tuning} \\
\hline
\multicolumn{1}{l}{ImageNet} & \multicolumn{1}{c}{Human} & \multicolumn{1}{c}{Classification (1,000, 1.2 million)} & \multicolumn{1}{c}{Person detection}\\
\multicolumn{1}{l}{ECP} & \multicolumn{1}{c}{Human} & \multicolumn{1}{c}{Person detection (2, 240,000)} & \multicolumn{1}{c}{Person detection}\\
\multicolumn{1}{l}{Pascal VOC} & \multicolumn{1}{c}{Human} & \multicolumn{1}{c}{Object detection (20, 10,000)} & \multicolumn{1}{c}{Person detection}\\
\multicolumn{1}{l}{WSPD (ours)} & \multicolumn{1}{c}{Weak} & \multicolumn{1}{c}{Person detection (2, 8.7 million)} & \multicolumn{1}{c}{Person detection} \\
\hline
\end{tabular}
\end{center}
\vspace{0pt}\caption{Annotation type, pre-training, and fine-tuning for each method.}
\label{tab:pretrainfinetune} \end{table*}

\section{Configuration for Detectors}
In this section, we describe a suitable base model and training configuration. 

\subsection{Representative architecture in person detection}
To assess the performance achieved when using the proposed WSPD collection method, we consider different types of representative detectors, namely the SSD~\cite{refs:SSD} and M2Det~\cite{refs:M2Det}. In our explorative analysis, we utilized the WSPD dataset containing over 8.7 million bboxes to optimize the network parameters for bbox regression and person classification. The hyperparameter settings used here were the same  as those in~\cite{refs:SSD,refs:M2Det}.




\begin{table*}
\begin{center}
\begin{tabular}{lllcc}
\hline
\multicolumn{1}{c}{Method} &
\multicolumn{1}{c}{Pre-training} &
\multicolumn{1}{c}{Pre-training} & \multicolumn{1}{c}{\#batches, \#epochs} &  \multicolumn{1}{c}{Miss rate (\%)}\\
\multicolumn{1}{l}{} & \multicolumn{1}{l}{} &  \multicolumn{1}{c}{supervision} & \multicolumn{1}{c}{} & \multicolumn{1}{c}{(lower is better)}\\
\hline
\multicolumn{1}{l}{SSD} &  \multicolumn{1}{l}{ImageNet} & \multicolumn{1}{l}{Human} & \multicolumn{1}{c}{64, 100} & \multicolumn{1}{c}{33.90}\\
\multicolumn{1}{l}{SSD} &  \multicolumn{1}{l}{Pascal VOC} & \multicolumn{1}{l}{Human} & \multicolumn{1}{c}{64, 100} & \multicolumn{1}{c}{29.28}\\
\multicolumn{1}{l}{SSD} &  \multicolumn{1}{l}{ECP} & \multicolumn{1}{l}{Human} & \multicolumn{1}{c}{64, 100} & \multicolumn{1}{c}{26.92}\\

\multicolumn{1}{l}{M2Det320} & \multicolumn{1}{l}{ImageNet} & \multicolumn{1}{l}{Human} & \multicolumn{1}{c}{16, 100} & \multicolumn{1}{c}{57.31}\\
\multicolumn{1}{l}{M2Det320} &  \multicolumn{1}{l}{Pascal VOC} & \multicolumn{1}{l}{Human} & \multicolumn{1}{c}{16, 100} & \multicolumn{1}{c}{73.72}\\
\multicolumn{1}{l}{M2Det320} &  \multicolumn{1}{l}{ECP} & \multicolumn{1}{l}{Human} & \multicolumn{1}{c}{16, 100} & \multicolumn{1}{c}{97.68}\\


\multicolumn{1}{l}{M2Det512} &  \multicolumn{1}{l}{ImageNet} & \multicolumn{1}{l}{Human} & \multicolumn{1}{c}{8, 100} & \multicolumn{1}{c}{32.46}\\
\multicolumn{1}{l}{M2Det512} &  \multicolumn{1}{l}{Pascal VOC} & \multicolumn{1}{l}{Human} & \multicolumn{1}{c}{8, 100} & \multicolumn{1}{c}{23.05}\\
\multicolumn{1}{l}{M2Det512} &  \multicolumn{1}{l}{ECP} & \multicolumn{1}{l}{Human} & \multicolumn{1}{c}{8, 100} & \multicolumn{1}{c}{82.53}\\

\hline \hline

\multicolumn{1}{l}{SSD (ours)} &  \multicolumn{1}{l}{WSPD} & \multicolumn{1}{l}{Weak} & \multicolumn{1}{c}{128, 25} & \multicolumn{1}{c}{24.06}\\
\multicolumn{1}{l}{SSD (ours)} &  \multicolumn{1}{l}{WSPD} & \multicolumn{1}{l}{Weak} & \multicolumn{1}{c}{128, 50} & \multicolumn{1}{c}{20.95}\\
\multicolumn{1}{l}{SSD (ours)} &  \multicolumn{1}{l}{WSPD} & \multicolumn{1}{l}{Weak} & \multicolumn{1}{c}{128, 100} & \multicolumn{1}{c}{\textbf{20.55}}\\
\multicolumn{1}{l}{SSD (ours)} &  \multicolumn{1}{l}{WSPD} & \multicolumn{1}{l}{Weak} & \multicolumn{1}{c}{256, 25} & \multicolumn{1}{c}{24.35}\\
\multicolumn{1}{l}{SSD (ours)} &  \multicolumn{1}{l}{WSPD} & \multicolumn{1}{l}{Weak} & \multicolumn{1}{c}{256, 50} & \multicolumn{1}{c}{22.92}\\
\multicolumn{1}{l}{SSD (ours)} &  \multicolumn{1}{l}{WSPD} & \multicolumn{1}{l}{Weak} & \multicolumn{1}{c}{256, 100} & \multicolumn{1}{c}{21.45}\\

\multicolumn{1}{l}{M2Det320 (ours)} &  \multicolumn{1}{l}{WSPD} & \multicolumn{1}{l}{Weak} & \multicolumn{1}{c}{16, 50} & \multicolumn{1}{c}{\textbf{16.44}}\\
\multicolumn{1}{l}{M2Det512 (ours)} &  \multicolumn{1}{l}{WSPD} & \multicolumn{1}{l}{Weak} & \multicolumn{1}{c}{8, 35} & \multicolumn{1}{c}{18.85}\\






\hline
\end{tabular}
\end{center}
\vspace{0pt}\caption{Detection performance comparisons for the Caltech Pedestrian. We list the method, backbone network, pre-trained dataset, supervision during pre-training, size of batch, number of pre-training epochs, and miss rate (\%). Though our WSPD applies only weak supervision during pre-training, we achieve higher rates on fine-tuning tasks.}
\label{tab:explore}
\end{table*}

\subsection{Training method for each dataset}

We conducted pre-training with our WSPD and fine-tuning for each person dataset. Throughout the experiment, we evaluated the pre-trained models; therefore, fine-tuning was conducted for all pre-trained models on the pedestrian dataset. Our WSPD pre-trained model is compared with three different models: ImageNet, Pascal VOC, and EuroCity Persons pre-trained detector. The ImageNet pre-trained model is trained with a large number of images, but no bboxes are used during the pre-training. In contrast, the EuroCity Person pre-trained detector uses 240,000 person bboxes in the pre-training step. Moreover, the Pascal VOC pre-trained detector is not limited to person bboxes, and it has 20 object annotations. We show the procedures used for pre-training and fine-tuning in Table~\ref{tab:pretrainfinetune}. We used the Caltech Pedestrian in the fine-tuning step.



\section{Experimental Results and Discussion}

This section clarifies how the use of the weakly supervised dataset collection influences the accuracy of a pre-trained person detector. The weak but numerous annotations enable us to improve the performance in the fine-tuning task. The resulting accuracy is higher than that of other fully supervised pre-trained models (Table~\ref{tab:pretrainfinetune}). We also present the results of the exploratory analysis and compare the performance when using different detection architectures.

\subsection{Exploratory study}

The purpose of the exploratory study was to optimize the  hyperparameters for each architecture using the self-assembled dataset. Although there are numerous hyperparameters that must be selected in the detection architecture and learning strategy, we examined the effects of batch size \{128, 256\} and \#epoch \{25, 50, 100\} with the WSPD method, as they seem to be the most important for model training. Therefore, we here calculate six different pre-trained SSD models in the pre-training phase. In addition to using the pre-trained detectors, we conducted further fine-tuning on a target dataset. To simplify the parameter tuning step, we employed the SSD (detection architecture), WSPD (pre-trained dataset), and Caltech Pedestrian (fine-tuned dataset).

The exploration results for the Caltech Pedestrian are shown in Table~\ref{tab:explore}. The table shows the change in the miss rates (lower is better) for different numbers of batches and pre-training epochs. According to the results, we can confirm that 128 batches and 100 pre-training epochs provide the best performance.  Additionally, we found that the number of pre-training epochs tends to perform better when the pre-training time is longer. However, a smaller  number of pre-training epochs must be considered because the training time with 8.7 million bboxes is high. Especially during pre-training with the WSPD, the average training time is roughly 39 hours per epoch on four NVIDIA Tesla V100 GPUs. Therefore, pre-training with 100 epochs requires approximately 3,969 hours (165 days). Undoubtedly, the longer training will result in better pre-training results, but we must consider a more reasonable training time on larger detection architectures like M2Det. 

\begin{figure*}
\begin{center}
\includegraphics[width=1.0\linewidth]{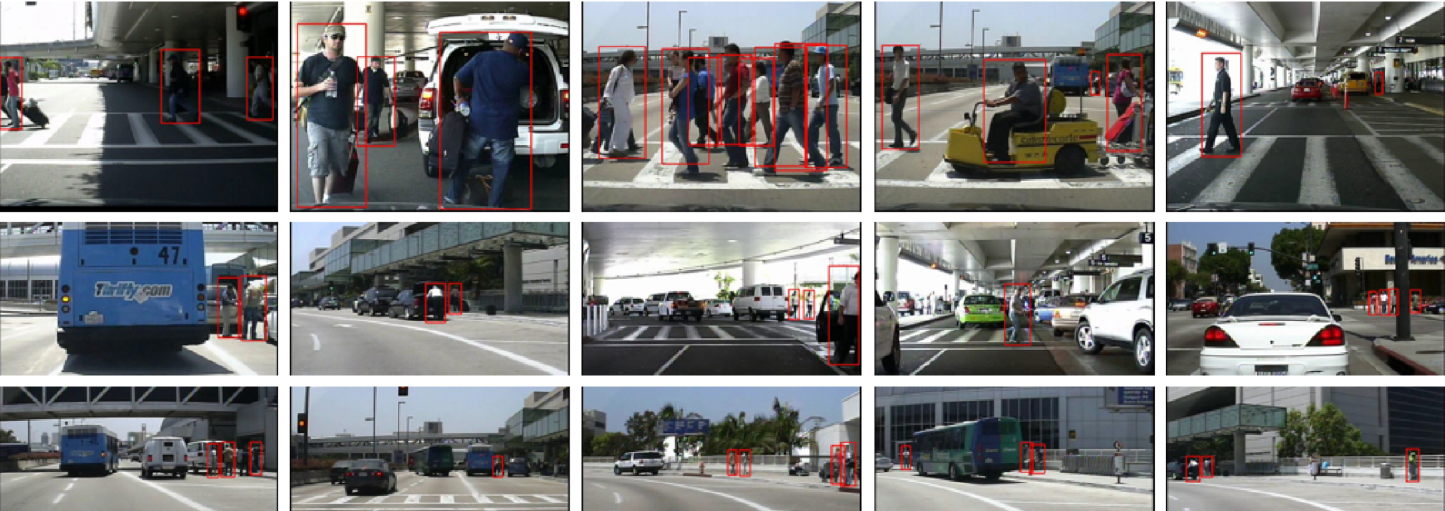}
\end{center}
   \vspace{0.0pt}\caption{\cHiro{Detection examples with WSPD pre-trained SSD.}}
\label{fig:example}
\end{figure*}

\begin{figure}[t]
\begin{center}
\includegraphics[width=1.0\linewidth]{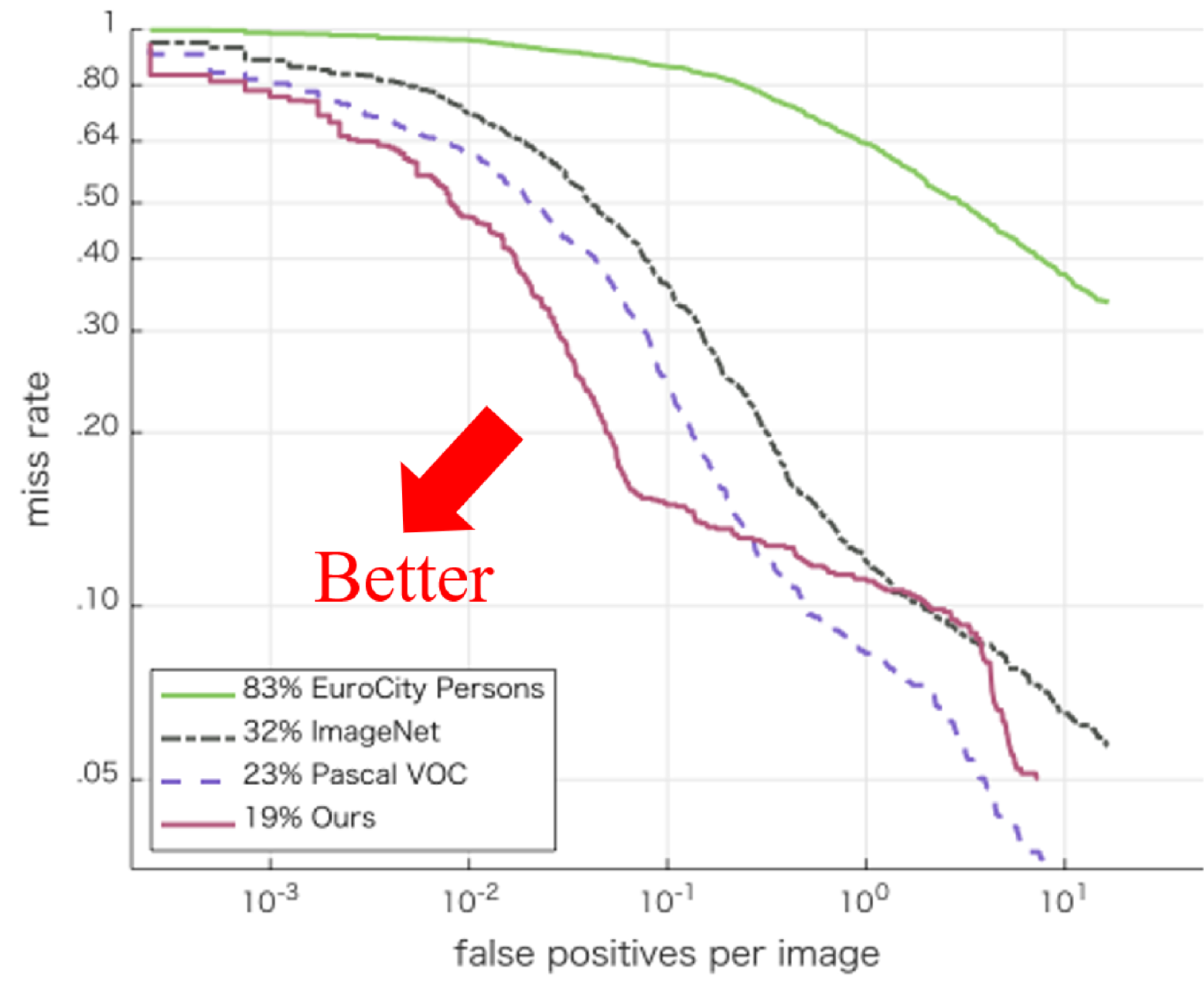}
\end{center}
   \vspace{0.0pt}\caption{DET curves for the M2Det512 model.}
\label{fig:m2det}
\end{figure}
\subsection{Comparison with baseline models}

We consider the detection results in detail for each architecture, backbone network, pre-trained dataset, and miss rate in Table~\ref{tab:explore}. We focus on the validation of the architectures (SSD/M2Det) and pre-trained datasets (ImageNet, Pascal VOC, EuroCity Persons, and WSPD). We discuss the results for both the SSD and M2Det architectures.

\textbf{SSD.} Figure~\ref{fig:results} shows the results for the proposed method and three baselines with the SSD architecture. The difference between our proposed method and the baselines for the pre-training tasks is shown in Table~\ref{tab:pretrainfinetune}. Basically, the pre-training results with the WSPD are significantly different because the dataset was collected in a weakly supervised manner. We confirmed that our WSPD pre-trained model achieved the highest score of 20.54\%, which is a 6.38\% and 13.36\% better miss rate than the models pre-trained on EuroCity Persons and ImageNet, respectively. Note that the weakly supervised dataset collection for person detection was processed by a two-step algorithm using an existing detector and binary classification. Despite the presence of noise in the dataset, our method outperformed the fully supervised bboxes developed by human annotators. This result suggests that we can automatically generate a ground truth dataset in a simple way. The performance rate is higher than for the Pascal VOC pre-trained model (our method has an 8.74\% better miss rate), which assigns multiple object detection labels in the pre-trained phase.

\textbf{M2Det.} In addition to the SSD model, we considered the M2Det (320/512) model. The M2Det detector represents the current state of the art in terms of detector accuracy. As described above, we compared the self-collected WSPD with ImageNet, Pascal VOC, and EuroCity Persons in the context of the pre-trained dataset with M2Det512 (see Figure~\ref{fig:m2det}). According to Table~\ref{tab:explore}, the best miss rate was 16.44\% with M2Det320. The miss rate is 4.01\% better than that for the WSPD pre-trained SSD. The ImageNet pre-trained M2Det512 had a 32.46\% miss rate on the Caltech Pedestrian. 

Moreover, we list the detection comparisons and results in Figure~\ref{fig:results} and Figure~\ref{fig:example}, respectively. 


\begin{figure}[t]
\begin{center}
\includegraphics[width=1.0\linewidth]{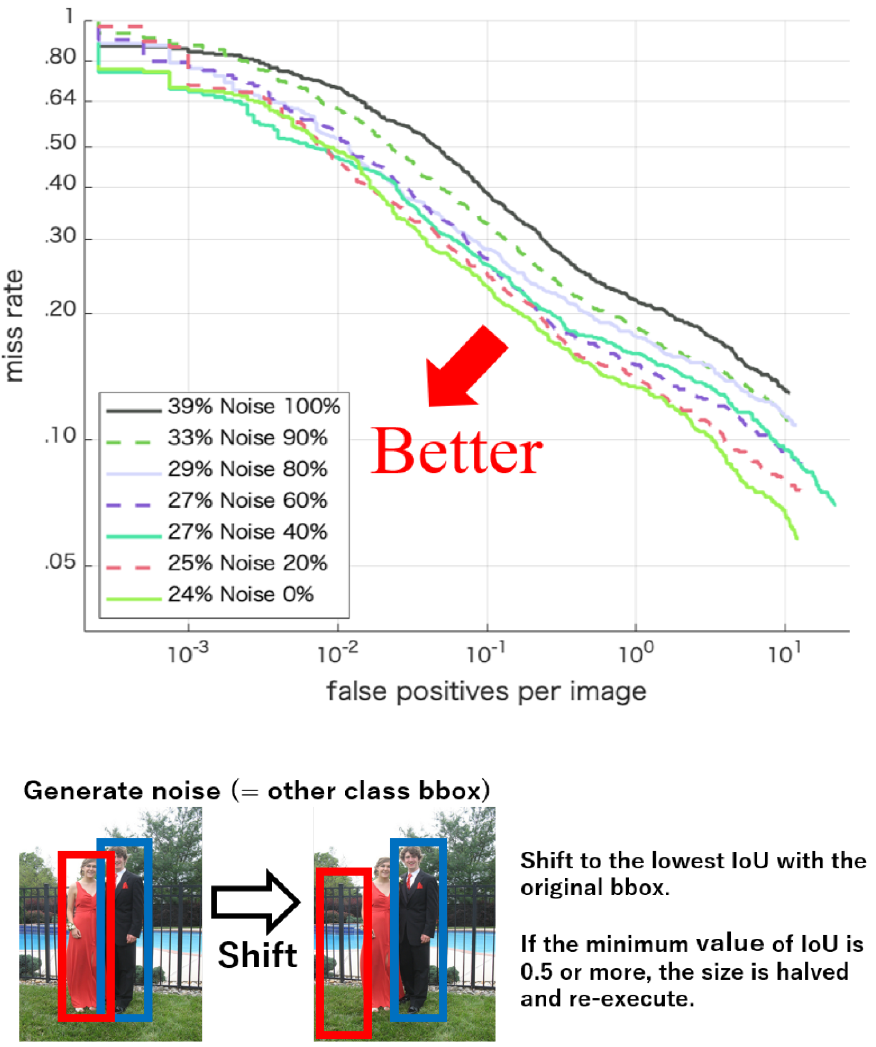}
\end{center}
   \vspace{0pt}\caption{(Top) Relationship between additional noise rate (0\%, 20\%, 40\%, 60\%, and 80\%) and detection miss rate. (Bottom) How to create a "noisy" bbox from an image.}
\label{fig:noiseanalysis}
\end{figure} 

\begin{table}
\begin{center}
\begin{tabular}{ccc}
\hline
\multicolumn{1}{c}{Noise} & \multicolumn{1}{c}{Miss rate} & \multicolumn{1}{c}{Difference from normal training}\\
\multicolumn{1}{c}{(\%)} & \multicolumn{1}{c}{(\%)} & \multicolumn{1}{c}{(\%)}\\\hline
\multicolumn{1}{c}{0} & \multicolumn{1}{c}{23.86} & \multicolumn{1}{c}{--}\\
\hline \multicolumn{1}{c}{10} & \multicolumn{1}{c}{24.06} & \multicolumn{1}{c}{-0.20}\\
\multicolumn{1}{c}{20} & \multicolumn{1}{c}{24.65} & \multicolumn{1}{c}{-0.79}\\
\multicolumn{1}{c}{30} & \multicolumn{1}{c}{24.81} & \multicolumn{1}{c}{-0.95}\\
\multicolumn{1}{c}{40} & \multicolumn{1}{c}{26.82} & \multicolumn{1}{c}{-2.96}\\
\multicolumn{1}{c}{50} & \multicolumn{1}{c}{25.49} & \multicolumn{1}{c}{-1.63}\\
\multicolumn{1}{c}{60} & \multicolumn{1}{c}{27.25} & \multicolumn{1}{c}{-3.39}\\
\multicolumn{1}{c}{70} & \multicolumn{1}{c}{29.68} & \multicolumn{1}{c}{-5.82}\\
\multicolumn{1}{c}{80} & \multicolumn{1}{c}{28.98} & \multicolumn{1}{c}{-5.12}\\
\multicolumn{1}{c}{90} & \multicolumn{1}{c}{33.03} & \multicolumn{1}{c}{-9.17}\\
\multicolumn{1}{c}{100} & \multicolumn{1}{c}{38.62} & \multicolumn{1}{c}{-14.76}\\
\hline
\end{tabular}
\end{center}
\vspace{0pt}\caption{Detailed noise rate and miss rate correspondences. We also show the difference from normal training, which has a miss rate of 23.86\%.}
\label{tab:noise}
\end{table}


\subsection{Noise label analysis}
Additionally, we investigated the effect of label noise.

In addition to the manual data quality analysis (see Table~\ref{tab:dataquality}), we analyzed the relationship between the amount of noise in the dataset and the detection accuracy. We deliberately added a bbox translation with horizontal and vertical movement in the ($x,y$) coordinates. The procedure of making noise is shown at the bottom of Figure~\ref{fig:noiseanalysis}. We translated a bbox in the image and prevented it from overlapping a ground truth bbox.  We simultaneously list the relationship between noise rate and miss rate in the top of Figure~\ref{fig:noiseanalysis} and in Table~\ref{tab:noise}. In the figure and table, 0\% noise (miss rate of 23.86\%) corresponds to normal training  and 100\% noise (miss rate of 38.62\%) corresponds to translating all bboxes. In the experiment, note that the data was obtained as 1 million bboxes randomly selected from the WSPD dataset; therefore, the miss rate is different from the 20.55\% with 8.7M bboxes shown in Table~\ref{tab:explore}.

According to the results, 30\% noise produced only a small increase in the miss rate (from 23.86 to 24.81\%, a difference of 0.95\%). This confirmed that a small amount of noise does not greatly affect the performance rate of person detection. At noise rates greater than 30\%, the miss rate continued to increase at 40\% (miss rate of 26.82\%, difference of 2.96\%) to 80\% noise (miss rate of 28.98\%, difference of 5.12\%). The 90 and 100\% noise rates produced the worst results, with differences of 9.17 and 14.76\% from the normal training rate. \cHiro{In particular, the results with the 100\% noise rate are worse than those with ImageNet pre-training with the Caltech Pedestrian.}

\section{Conclusion}
This paper proposes a weakly supervised dataset collection method for improving pre-trained person detection models. In a comparison with the baseline detectors (e.g., ImageNet pre-trained model and person dataset fine-tuned model), our proposed method achieved a 6.38 and 13.36\% better miss rate than the EuroCity Persons and ImageNet pre-trained models, respectively. The semi-automatic image and bbox collection can be performed by downloading images from SNS (e.g., Flickr), and using an existing detector (e.g., Faster R-CNN) for binary classification to determine whether an image contains a person's whole body. The weakly supervised dataset collection approach is simple yet highly effective for producing a pre-trained detector. We confirmed that using a large number of bboxes (8.7 million boxes in the WSPD dataset) in the pre-training task results in performance much better performance than that of the baseline detectors.

{\small
\bibliographystyle{ieee_fullname}
\bibliography{egpaper_final}
}

\end{document}